\pdfoutput=1
\documentclass[11pt]{article}

\usepackage[final]{acl}

\usepackage{times}
\usepackage{latexsym}
\usepackage{multirow}

\usepackage[T1]{fontenc}

\usepackage[utf8]{inputenc}

\usepackage{microtype}
\usepackage{ulem}
\usepackage{inconsolata}
\usepackage{booktabs}
\usepackage{graphicx}
\usepackage{framed}
\usepackage{array}
\usepackage{dashrule}
\usepackage{tikz}
\usetikzlibrary{fit}
\usepackage{subcaption}
\usepackage{CJK}
\usepackage{makecell}
\usepackage[most]{tcolorbox}
\usepackage{cleveref}

\newcommand{\figref}[1]{Figure \ref{#1}}

\newcommand{\KZ}[1]{\textcolor{blue}{(Kenny: #1})}

\definecolor{shadecolor}{RGB}{224, 224, 224}
\definecolor{framecolor}{RGB}{160,160,160}

\title{\uline{Mix}ed Chain-of-\uline{Psy}chotherapies for Emotional Support Chatbot}

\author{Siyuan Chen$^1$, Cong Ming$^{2}$, Zhiling Zhang$^{3}$, Yanyi Chen$^4$,  Kenny Q. Zhu$^{5}$, Mengyue Wu$^6$\thanks{~~Corresponding author.}  \\ 
$^{1,2,6}$
X-LANCE Lab, Department of Computer Science and Engineering\\
MoE Key Lab of Artificial Intelligence, AI Institute\\$^{1,2,3,6}$Shanghai Jiao Tong University, Shanghai, China \\
$^{4}$ Rutgers University, New Jersey, USA\\
$^{5}$University of Texas at Arlington, Arlington, Texas, USA\\
\texttt{\{$^1$chensiyuan925, $^6$mengyuewu\}@sjtu.edu.cn,}
}

\begin{document}
\maketitle
\begin{abstract}
In the realm of mental health support chatbots, it is vital to show empathy and encourage self-exploration to provide tailored solutions. However, current approaches tend to provide general insights or solutions without fully understanding the help-seeker's situation. Therefore, we propose \textit{PsyMix}, a chatbot that integrates the analyses of the seeker's state from the perspective of a psychotherapy approach (Chain-of-Psychotherapies, \textit{CoP}) before generating the response, and learns to incorporate the strength of various psychotherapies by finetuning on a mixture of CoPs. Through comprehensive evaluation, we found that \textit{PsyMix} can outperform the ChatGPT baseline, and demonstrate a comparable level of empathy in its responses to that of human counselors.
\end{abstract}

\section{Introduction}
\label{sec:intro}
Mental health plays a crucial role in personal well-being, yet there's a shortage of specialized mental healthcare resources worldwide~\cite{Sharma_Choudhury_Althoff_Sharma_2020}. To bridge this gap, AI-driven chatbots have emerged as promising solutions~\cite{welivita-pu-2023-boosting}. These chatbots function as virtual emotional supporter, offering convenient avenues for individuals to access support in their daily lives. 

\begin{figure}[th]
	\centering
	\includegraphics[width=0.8\linewidth]{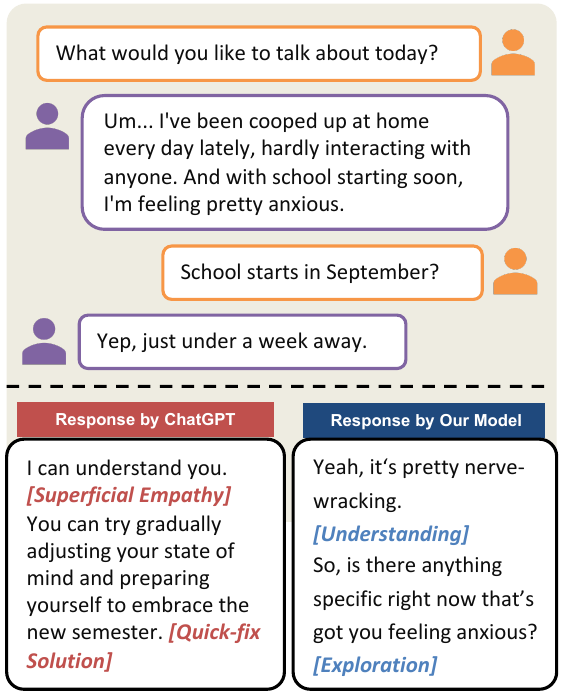}
	\caption{Comparison of the responses generated by ChatGPT and \textit{PsyMix} based on the same context.}
	\label{fig:overview}
\end{figure}

Numerous attempts have been made to develop emotional support chatbots, particularly in the era of Large Language Models (LLMs). However, current approaches~\cite{Qiu2023SMILEST, loh2023harnessing, Chen2023LLMempoweredCF} providing emotional support by prompting LLMs fall short on several dimensions.
The first one can be termed as ``\textit{\textbf{superficial empathy}}'', which manifests as vague and generalized responses aimed at conveying empathy (e.g., ``I can understand your feelings''). These responses lack profound and specific comprehension of the user's context, potentially leaving them feeling inadequately understood~\cite{lee-etal-2023-empathy, Sharma2021Empathy}.
Another issue is referred to as ``\textit{\textbf{Quick-fix Solution}}'', as LLMs tend to offer general solutions as soon as the user expresses concerns, without exploring the underlying reasons or less apparent issues. These quick fixes are usually insufficient to align with the nuanced requirements of emotional support~\cite{Sharma_Choudhury_Althoff_Sharma_2020}, where supporters should encourage self-exploration to provide tailored support~\cite{lahnala-etal-2022-critical}.

Both of these issues stem from a lack of comprehensive understanding and professional insight into seeker's situation through the application of psychotherapy knowledge. 
Drawing inspiration from real counseling scenarios, where seasoned counselors often blend various psychotherapy approaches to analyze the seeker's situation~\cite{lee2023chain}, we endeavor to integrate multiple mainstream therapy approaches into LLM tuning for emotional support. 
However, modeling after human counselors poses several challenges: 1) the internal analysis performed by a counselor is not explicitly conveyed in their responses, making it difficult to replicate this cognitive process; 2) there currently lacks datasets that differentiate counseling dialogues by psychotherapy approach. Moreover, since counselors often blend approaches, distinguishing between them may also be inherently unnatural. Consequently, there is an absence of dialog datasets with psychotherapy-related annotations.


Therefore, our work seeks to overcome these hurdles by introducing \textit{PsyMix}, a conversational agent designed for emotional support. To tackle the intricacy of integrating expert knowledge, we leverage ChatGPT to reconstruct the analysis of seekers' situations at the sentence level. 
With expert-directed dimensional guidelines, we generate professional analyses from the viewpoints of various psychotherapy approaches (such as cognitive behavioral therapy, CBT) for each response in existing pure-dialog datasets provided by human counselors~\cite{li-etal-2023-understanding}, and these analyses are denoted as \textbf{\textit{Chain-of-Psychotherapies (CoP)}}. 
Subsequently, we employ supervised fine-tuning (SFT) to tune an LLM using data featuring mixed analyses from diverse psychotherapies. These analyses guide the response generation process, enhancing the agent's ability to deliver tailored support. Experimental results demonstrate the superiority of our approach with CoP over naive SFT methods. Additionally, the integration of mixed CoP in the tuning data enables insights from diverse analytical perspectives, further improving performance over single-aspect CoP methods.

\section{Psychotherapy Approaches}
\label{sec:psychotherapy}

Given that our solution centers on psychotherapy approaches, we first provide a brief overview of the involved approaches in this work, which are highly representative and widely utilized in practice.

\textbf{Cognitive Behavioral Therapy (CBT)} is a structured and goal-oriented psychotherapy that empowers individuals to become their own therapists. It focuses on helping them recognize their thoughts and behaviors, providing skills to change maladaptive cognitive and behavioral patterns~\cite{Kristina2013KeyCBT}. To achieve these goals, the ABC model was developed, which suggests that ``our emotions and behaviors (C: Consequences)'' are not directly determined by life events (A: Activating Events), but rather by the way these events are cognitively processed and evaluated (B: Beliefs)''~\cite{Oltean2017AnEA}.

\textbf{Person Centered Therapy (PCT)} is a human-based therapy with the belief that the seeker is innately motivated and has the capacity for growth and self-actualization~\cite{Rogers1946SignificantAO}. It is a supportive, positive, and nondirective therapy that aims to provide seekers with a non-judgmental environment that helps them explore themselves honestly and thoroughly~\cite{Elliott2013PCT}. The therapist encourages seekers to explore their emotions and experiences in order to discover their inner strength and gain self-awareness.

\textbf{Solution-Focused Brief Therapy (SFBT)} is a competency-based model that prioritizes seekers' strengths over past failures. It facilitates the envisioning of a future where problems are resolved, constructing a pathway collaboratively with seekers. This pathway is tailored to their individual goals, resources, actions, etc.~\cite{Trepper2014SolutionFocusedTT}.

\section{Method}
\label{sec:method}

In this section, we introduce the two-step pipeline (Figure \ref{fig:prompt_resp}) of \textit{PsyMix} to fine-tune trainable LLM for better generation.

\subsection{Chain-of-Psychotherapies (CoP)}
\label{sec:cot_generation}
\begin{figure}[th]
	\centering
	\includegraphics[width=\linewidth]{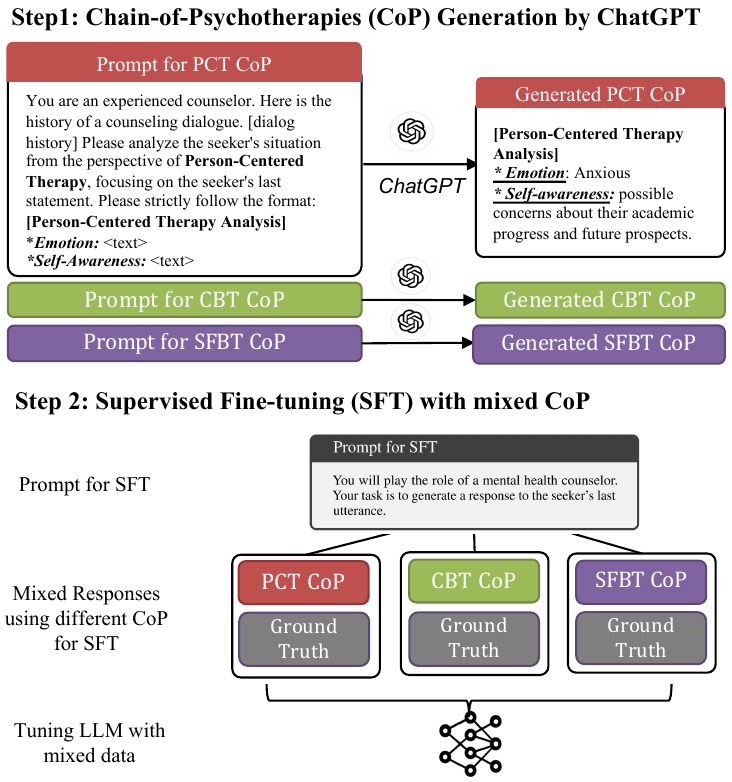}
	\caption{Illustration of the two-step pipeline. We first generate CoPs via ChatGPT, then pack CoP with the original utterance (i.e., ground truth) in the dialog for supervised fine-tuning.}
	\label{fig:prompt_resp}
\end{figure}

As Section \ref{sec:intro} mentions, human counselors always analyze the seeker's situation with the perspective of various psychotherapy approaches before providing responses. However, these analyses are not explicitly manifested in dialogues. Therefore, our first step involves generating such analysis with ChatGPT, and we refer to these analyses as \textbf{\textit{Chain-of-Psychotherapies} (CoP)} (Step 1 in Figure \ref{fig:prompt_resp}).

\begin{table}[th]
    \small
    \centering
    \begin{tabular}{c|c}
    \toprule
    Psychotherapy & Perspective of Thought  \\ 
    \midrule
    CBT &  Event, Cognition, Behavior, Belief \\
    PCT &  Emotion, Self-Awareness \\
    SFBT & Goal, Resource, Exception, Action \\
    \bottomrule
    \end{tabular}
    \caption{Different Psychotherapy Thoughts}
    \label{tab:psychotherapy_thoughts}
\end{table}

We employ three widely influential psychotherapy approaches, as detailed in Section \ref{sec:psychotherapy}. To ensure the professionalism and reliability of the generated analyses, we do not allow ChatGPT to generate them independently. Instead, we collaborate with expert psychotherapists to define the specific dimensions that each psychotherapy approach should assess. These dimensions, outlined in Table \ref{tab:psychotherapy_thoughts}, represent the focal points of each approach. 
In the prompts, we instruct ChatGPT to generate analyses for every utterance made by seekers, covering all dimensions relevant to each psychotherapy approach. Detailed prompts to generate all the three CoPs are provided in Appendix \ref{apd:thought_prompt}.

\subsection{SFT with Mixed CoP}
\label{sec:mix}
In this step, we utilize the CoP generated by ChatGPT for Supervised Fine-Tuning (SFT).
For dialog context $c$, Chain-of-Psychotherapy $p$, and the corresponding ground truth response $r$, one approach to constructing the tuning (prompt, response) pair is to set $prompt=(c, p)$ and $response = r$. However, this approach proves ineffective in inference, as in practical usage, we still need to continuously call ChatGPT's API to generate CoP. Luckily, previous study~\cite{magister-etal-2023-teaching} shows that LLMs (e.g., ChatGPT) can impart the capabilities of CoT to smaller models through knowledge distillation. Therefore, we opt to teach smaller models to generate CoP on their own, and set $prompt=c$, $response = (p, r)$, as illustrated in \figref{fig:prompt_resp}. 

Moreover, inspired by the mixed thinking process in real counselling sessions, we utilize mixed Chain-of-Psychotherapies, which mixed analyses derived from all the three psychotherapy approaches for LLM-tuning. The training loss is: $L_{mix} = -\sum^3_{i=1} \log P(p_i, r| c)$,
where $c$ is the context, $p_i$ is the CoP of the $i$-th psychotherapy, and $r$ is the ground truth response.
This methodology not only encourages the model to ``think'' before generating response, but also integrates strengths from various analytical perspectives. Then in inference, the model will generate $p$, where the applied psychotherapy is decided, along with $r$.

\section{Experiments}
\label{sec:exp}

We demonstrated the effectiveness of our \textit{PsyMix} in this section through human evaluation and automatic analysis through GPT-4~\cite{openai2023gpt4}. 
\subsection{Dataset}
The dataset for supervised fine-tuning, named \texttt{Xinling}~\cite{li-etal-2023-understanding}, is a public real counseling dialogue dataset in Chinese. This dataset comprises 300 dialogues between counselors and seekers, gathered from an online welfare counseling platform.  Each session on this platform provides about 50 minutes (with an average of 78 turns) of free text-based counseling delivered by experienced counselors, highlighting the dataset's high quality.
\subsection{Experiment Settings}
The base model utilized for supervised fine-tuning is \texttt{Baichuan2-7B(Chat)}~\cite{yang2023baichuan}, currently one of the top-performing Chinese LLMs of moderate size. Nevertheless, our methodologies are also applicable to models and datasets in other languages. The hyper-parameters employed in tuning are detailed in Appendix  \ref{apd:experiment}.

\subsection{Methods for Comparison}
\label{sec:compare}
In our experiments, we primarily compare our proposed \textit{PsyMix} with two baselines and single-aspect CoP. 
Baselines includes \textbf{Naive}, which involves straightforward supervised fine-tuning without incorporating CoP, and  \textbf{ChatGPT}, referring to the responses generated directly by prompting ChatGPT as a counselor\footnote{We provide the prompt in Appendix \ref{apd:baseline_prompt}}. 
We also trained models exclusively on one psychotherapy approach's analysis, respectively named \textbf{PCT CoP}, \textbf{CBT CoP} and \textbf{SFBT CoP}.

\subsection{Human Evaluation}
Human evaluation is widely considered the golden standard for dialog systems, particularly for counseling sessions where subjective satisfaction is the optimal goal. We recruited evaluators through online advertisements, resulting in 16 volunteers aged 19 to 51, 56\% of which were female.
\paragraph{Evaluation Procedure} 
For counseling dialogues, immersion in the conversation is crucial for accurate assessment. Evaluators must fully engage to better understand the seeker's needs and assign more reasonable scores. Additionally, counseling sessions tend to be lengthy. Therefore, our chosen method of human evaluation involves having each evaluator rate an entire counseling dialogue sentence by sentence, from start to finish. This approach is efficient and enables evaluators to immerse themselves in the conversation. 
For each utterance from the seeker, we provide responses from the seven different models of counselors outlined in Section \ref{sec:compare} in a randomized order. Evaluators then assign scores from 1 to 5 to each response, based on how comfortable they feel and inclined to continue the coversation. 

\begin{table}[th]
    \centering
    \small
    \resizebox{\columnwidth}{!}{
    \begin{tabular}{lccc}
    \toprule
     Method & Avg. score & Avg. length & Satisfaction rate \\ 
    \midrule
    \textit{ChatGPT}  &  2.73 & 47.06 &  33.9\%   \\ 
    \textit{naive}     &   2.81   & 14.56 & 32.1\%\\
    \midrule
    \textit{PsyMix}  &   \uline{3.12}  & 25.97 & \uline{40.8\%}  \\ 
    \textit{CBT CoP}  &  3.04  & 24.91 & 38.6\%   \\
    \textit{PCT CoP}  &  2.99  & 25.67  & 37.5\%  \\ 
     \textit{SFBT CoP}  &  2.96  & 24.93 & 37.6\%\\
     \midrule
     \textit{ground truth} & \textbf{3.58} & 25.72 & \textbf{55.5\%}   \\
    \bottomrule
    \end{tabular}
    }
    \caption{Human-rated scores of responses generated by various counselor chatbots and human counselors. ``Satisfaction rate'' means the proportion of utterances scored 4 or 5 points by humans.}
    \label{tab:human}
\end{table}

\paragraph{Results}
We assessed a total of 348 utterances extracted from 12 dialogues, with each utterance rated by two evaluators. The average pairwise agreement\footnote{This metric is introduced in~\citet{liu2023alignbench}. The detailed description is in Appendix \ref{apd:agreement}.} among these evaluators is 0.723, indicating a relatively high level of agreement. The rating results are in Table \ref{tab:human}. 
We observed that the ground truth responses produced by human counselors achieved the highest average score. Among the responses generated with CoP, \textit{PsyMix}, which incorporates mixed CoP of three psychotherapy approaches, outperformed others that used only one type of CoP. Interestingly, we noticed that incorporating CoP led to a significant increase in response length and higher scores, compared to the naive approach. This improvement is because the greater specificity of CoP-enhanced responses by deeply understanding the seeker's situation\footnote{We provide specific examples generated by these models as case studies in Appendix \ref{apd:case}}.

\subsection{Empathy Analysis}
\label{sec:empathy}
To analyze the characteristics of responses generated by different models, we utilize an empathy measure framework proposed by \citet{sharma-etal-2020-computational}. This framework dissects empathy into three dimensions: \textit{Emotional Reactions}, \textit{Interpretations}, and \textit{Explorations}. We employ GPT-4 to assign scores (1-3) to each aspect of every response. Detailed explanations of these three dimensions and scoring criteria are provided in Appendix \ref{apd:thought_prompt}.
\paragraph{Results} 
Human counselors' responses are regarded as ground-truth hence the more human-like, the better counselor chatbot is.
We therefore measure the empathy level of each generated response by calculating the Mean Square Error (MSE) between the response and the ground truth empathy scores. The results are displayed in Table \ref{tab:gpt4score-mse}. 
Notably, our model, \textit{PsyMix}, closely aligns with the ground truth across most dimensions, demonstrating an appropriate level of empathy.

\begin{table}[ht]
\centering
\resizebox{0.8\columnwidth}{!}{%
\small
\begin{tabular}{lcccc}
\toprule
\textbf{}             & ER & IP & EX & Average \\ 
\midrule
\textit{ChatGPT}      &0.966       &1.019     &1.557     &1.181\\
\textit{naive}        &\textbf{0.648}       &1.073     &1.461     &1.061\\
\midrule
\textit{PsyMix}    &0.725       &\textbf{0.902}     &\textbf{0.947}     &\textbf{0.858}\\
\textit{PCT CoP}      &0.761       & 0.952     &1.1101     &0.941\\ 
\textit{CBT CoP}      &0.668       &0.961     &1.1749     &0.935\\
\textit{SFBT CoP}     &0.757       &1.103     &1.102     &0.987\\ 
\bottomrule
\end{tabular}%
}
\caption{The Mean Square Error (relative to the ground truth, the lower the better) of GPT4-rated scores for various counselor chatbots. ER = Emotional Reaction, IP = Interpretation, EX = Exploration.}
\label{tab:gpt4score-mse}
\end{table}

\subsection{Case Study}
\label{apd:case}
After comparing the generation results of different models with the ground truth, we found that ChatGPT tends to generate more superficial empathy, often providing lengthy solutions and platitudes, but lacking in-depth probing questions. At times, it may appear rather abrupt and judgmental, conflicting with the principles advocated by modern psychological counseling. Meanwhile, responses generated by the naive model tend to be overly brief, lacking richness in expression and in depth empathy.

In contrast, \textit{PsyMix} performs closest to the ground truth, effectively balancing the proportions of various empathy dimensions, increasing the patient's desire to continue the conversation while expressing support and understanding. Although ChatGPT scored higher on average in empathy analysis, it performed poorly in MSE score and human evaluations, as shown in Tables \ref{tab:gpt4score-mse}, Table \ref{tab:gpt4score-origin}, and Table \ref{tab:human}. 

To provide a clearer and more intuitive comparison of the performance of different models, we offer some case analyses. For each case, we provide dialogue history and responses from \textit{PsyMix}, naive, ground truth, and ChatGPT, with empathetic components (\textcolor{orange!50!brown}{Emotional Reactions}, \textcolor{blue!50!black}{Interpretations}, and \textcolor{green!50!black}{Explorations}) in the responses marked in different colors.

In the examples shown in Table \ref{tab:case_anapysis_4}, the seeker and counselor discuss issues related to career development. After the seeker talked about their strengths at work, all four responses provided acknowledgment and support. However, the naive response lacked richer content, and the ChatGPT response was overly exaggerated and verbose. Like the ground truth, \textit{PsyMix} also affirmed the seeker while posing deeper questions, guiding them to further contemplate and express themselves.

As illustrated in the examples in Table \ref{tab:case_anapysis_2}, after the seeker expressed his/her view on university life, both the naive and \textit{PsyMix} models, like the ground truth, chose to further explore the seeker's perspective, with the ground truth posing more specific questions. Although ChatGPT expressed support and understanding, it failed to ask probing questions and instead offered lengthy opinions blindly, potentially causing the seeker to lose the desire to express themselves further.

In the examples shown in Table \ref{tab:case_anapysis_3}, the seeker recounted feeling guilt during childhood due to being sickly. The naive model's response simply reiterated what the seeker said. \textit{PsyMix} and the ground truth both asked deeper questions, with the ground truth attempting to explore the causes of the trauma. However, ChatGPT merely offered empty platitudes, potentially hindering further expression from the patient.

\begin{table*}[htbp]
    \centering
    \begin{tabular}{|p{15.5cm}|}
    \hline
    \textbf{Dialog History} 
    \par \textit{Seeker}: Yes. I'm not afraid of hardship or fatigue. As long as my goals are clear, I'm very resilient. What I fear most is not knowing what I want.
    \par \textit{Counselor}: Hmm, that's a significant advantage. What do you feel confident doing?
    \par \textit{Seeker}: I feel confident in interacting with people, helping others, making connections. I'm good at brainstorming, have a sharp mind, and strong communication skills.
    \\
    \hline
    \textbf{Response generation} 
    \par \textit{naive}: \textcolor{orange!50!brown}{Hmm, it sounds like a great ability.}
    \par \textit{gpt}: \textcolor{orange!50!brown}{Your observational skills and communication abilities must be excellent,} and these are very important strengths. Your natural talent for helping others and networking will definitely play a significant role in your future career development.
    \par \textit{PsyMix}: \textcolor{orange!50!brown}{Hmm, it sounds like a great ability.} \textcolor{green!50!black}{So, what significance do these abilities have for you?}
    \par \textit{ground truth}: \textcolor{orange!50!brown}{Hmm, that's great. } \textcolor{blue!50!black}{It seems like communication and networking are your strengths.}\textcolor{green!50!black}{ How can these strengths help you?}

    \\
    \hline
    \end{tabular}
    \caption{Case Analysis 1.} 
    \label{tab:case_anapysis_4}
\end{table*}

\begin{table*}[htbp]
    \centering
    \begin{tabular}{|p{15.5cm}|}
    \hline
    \textbf{Dialog History} 
    \par \textit{Counselor}: Hmm, have you thought about future career paths?
    \par \textit{Seeker}: Um, after going to college, I've thought about developing towards becoming a nutritionist.
    \par \textit{Counselor}: And now?
    \par \textit{Seeker}: I still have some thoughts now, but I'm afraid I won't be good enough because I feel like I don't understand the industry trends, and also I feel like college makes it easy for people to become demotivated.
    \\
    \hline
    \textbf{Response generation} 
    \par \textit{naive}: \textcolor{green!50!black}{Hmm, in what way?}
    \par \textit{gpt}: \textcolor{orange!50!brown}{I understand your concerns [Superficial Empathy]}, \textcolor{blue!50!black}{college life indeed presents some challenges, but it's also a process of growth.} You can try to learn more about the field of nutritionist, or participate in internships or activities related to it, gradually accumulating experience and understanding industry trends. As for feeling demotivated, you can also try to find some hobbies or join some clubs and activities to make your college life more fulfilling and meaningful\textcolor{red!50!black}{[Quick-fix Solution]}.
    \par \textit{PsyMix}: \textcolor{green!50!black}{Hmm, how do you understand it?}
    \par \textit{ground truth}: \textcolor{green!50!black}{Hmm, "college makes it easy for people to become demotivated," how do you mean?}
    \\
    \hline
    \end{tabular}
    \caption{Case Analysis 2.} 
    \label{tab:case_anapysis_2}
\end{table*}

\begin{table*}[htbp]
    \centering
    \begin{tabular}{|p{15.5cm}|}
    \hline
    \textbf{Dialog History} 
    \par \textit{Counselor}: How has your overall health been from childhood to now?
    \par \textit{Seeker}: Since I was little, I've had some digestive issues, and in [high school], I often caught colds. Once, it was so severe that I got sinusitis, and it still flares up from time to time. Also, my skin isn't great, and I have high degrees of myopia.
    \par \textit{Counselor}: Hmm, being ill can indeed affect the efficiency of other aspects of life.
    \par \textit{Seeker}: I envy those who are physically healthy, with normal height, normal vision, and rarely get sick. When I was little, whenever I got sick, I would always wonder if I had done something wrong and was being punished by a higher power.
    \\
    \hline
    \textbf{Response generation} 
    \par \textit{naive}: \textcolor{blue!50!black}{Hmm, it sounds like getting sick feels like a punishment.}
    \par \textit{gpt}: Everyone has their own bodily characteristics and health issues. It's important to learn to accept and take care of oneself. Instead of overly envying others, focusing on your own bodily needs is the most important thing\textcolor{red!50!black}{[Quick-fix Solution]}.
    \par \textit{PsyMix}: \textcolor{green!50!black}{Did you have such thoughts when you were sick as a child?}
    \par \textit{ground truth}: \textcolor{green!50!black}{Did you feel like you were wrong when you were sick as a child? Any specific incidents you associate with?}
    \\
    \hline
    \end{tabular}
    \caption{Case Analysis 3.} 
    \label{tab:case_anapysis_3}
\end{table*}

\section{Related Works}
\label{sec:related}

Empathy holds significant importance in conversations, particularly in scenarios involving mental health support~\cite{liu-etal-2021-towards}. 
Some approaches~\cite{Saha2022Motivational,zheng-etal-2023-augesc} focus on implementing counseling strategies, such as Hill's Helping skill~\cite{Hill2009Helping} or strategies from Motivational Interviewing~\cite{Miller1995MI}, to make responses appear more empathetic. However, these strategies often fall short in fully grasping the client's situation. This phenomenon has been highlighted in previous study~\cite{lee-etal-2023-empathy, Sharma2021Empathy}. Merely using patterned statements like ``I can understand your feelings'', regardless of context, can result in high scores in current empathy benchmarks~\cite{sharma-etal-2020-computational}, which is not reflective of genuine empathy.
Some prior approaches have started to pay attention to understanding the client's perspective. They commonly propose a taxonomy of user behavior, where before generating responses, the dialog system identifies and categorizes the user's emotions~\cite{rashkin-etal-2019-towards}, behaviors~\cite{qiu2023psychat}, or intentions~\cite{Su2023EmpatheticDG}. Moreover, some works~\cite{Li2020KnowledgeBF} integrate external common-sense knowledge to better comprehend clients. However, this approach typically tackles only one aspect, especially emotion, and the categorization outcomes are bound by the taxonomy or knowledge graph, thus restricting scalability.
\section{Conclusion}
\label{sec:conclusion}

In this work, we introduce \textit{PsyMix} to address the issue of ``superficial empathy'' and ``quick-fix solution'' of common counselling chatbots. The analysis of the seeker’s situation from various psychotherapy approaches (\textit{CoP}) are generated with the guideline from expert psychotherapist and further mixed for the finetuning of LLM. Human evaluations demonstrate that CoP-guided model surpasses ChatGPT baseline, and \textit{PsyMix} further improves upon any single CoP. Notably, \textit{PsyMix} delivers more specific and contextually relevant responses, with its empathy level closely matching human responses.

\section*{Ethical Considerations}
The dataset we use is from the study introduced by \citet{{li-etal-2023-understanding}}, which is granted ethics approval from the Institutional Ethics Committee (20211013LZZ001). We obtain the dataset through proper channels and ensure its use solely for this project's research. Bulk in-depth analysis of the dataset in this study is conducted by LLM and does not involve human annotators.

During manual evaluations, we fully consider diversity, with evaluators coming from different genders, ages, educational, and occupational backgrounds. They understand the significance of this study and willingly undertake unpaid evaluation work. We ensure that evaluators are mentally and physically healthy adults, and conduct follow-ups after evaluations to ensure they are not adversely affected by reading conversations with negative emotions.
\section*{Limitations}

In our study, there are some limitations that could be addressed in future research.
\begin{enumerate}
\item We currently only incorporate three mainstream psychotherapy approaches (i.e., PCT, CBT, SFBT) into Chain-of-Psychotherapies in this work, future explorations can include a wider range of psychotherapies like psychodynamics~\cite{Sheppard2006psychodynamics} to evaluate whether the more mixture, the better performance a chatbot will have.
\item  Despite our best efforts in designing the prompt for empathy analysis, the scores produced by GPT-4 are not highly accurate. Although GPT-4 is the most commonly employed model for assessing the generation effectiveness of LLMs~\cite{liu-etal-2023-g}, it tends to favor sentences generated by models like ChatGPT as higher quality, often deeming human responses as lower quality due to perceived inferior coherence compared to model-generated responses. Hence, it is imperative for us to explore more effective automated evaluation methods in the future.
\end{enumerate}

\bibliography{acl_latex}

\appendix

\section{Details of Experiment}
\label{apd:experiment}

\subsection{Experiment Settings}
We utilized OpenAI's \texttt{GPT-3.5-turbo} model to analyse seeker's state, setting the temperature to 0.1. 

The base model used for supervised fine-tuning is \texttt{Baichuan2-Chat 7B}, the hyper-parameters employed are as follows: the model is fine-tuned for 10 epochs using AdamW with a weight decay of 0.1, $\beta_1$ = 0.9, $\beta_2$ = 0.98, and $\epsilon = 1e^{-8}$. The learning rate was set to 2e-5, with a batch size of 8. 
The maximum context window length was limited to 4096 tokens and longer context are trimmed accordingly to fit within this constraint.
All checkpoints utilized in the final experiments correspond to epoch 5.

\subsection{Pairwise Agreement}
\label{apd:agreement}
Pairwise Agreement is a metric to measure the agreement between human evaluators~\cite{liu2023alignbench}. For each utterance, the human-judge scores were converted into pairwise comparisons, with $N$ responses yielding $C_n^2$ pairs. The agreement rate was then calculated across all pairs.  

\subsection{Significant Test for Human Evaluation}
\label{sec:ttest}
Table \ref{tab:ttest} shows the significance test results for human evaluation. It is evident from the table that the scores of \textit{ground truth} are significantly higher than all model-generated results, while the quality of \textit{PsyMix}'s generation is significantly better than all models except for CBT CoP. The differences among the three models exclusively on one psychotherapy approach's analysis are not pronounced, but they are all significantly better than the results generated by the naive and ChatGPT models. There is no significant difference between the naive and ChatGPT models.

\begin{table}[th]
    \centering
    \begin{tabular}{ll}
    \toprule
      Comparing results   & t-test \\   
    \midrule
    \textit{PsyMix} VS \textit{ground truth}    & t = $-6.50^{***}$     \\
    \textit{PsyMix} VS. \textit{PCT CoP}    & t = $1.83^{*}$  \\
    \textit{PsyMix} VS. \textit{CBT CoP}    & t = $1.19$  \\
    \textit{PsyMix} VS. \textit{SFBT CoP}    & t = $2.29^{**}$  \\
    \textit{PsyMix} VS. \textit{naive}    & t = $4.66^{***}$  \\
    \textit{PsyMix} VS. \textit{ChatGPT}    & t = $4.78^{***}$  \\
    \midrule
    \textit{CBT CoP} VS \textit{ground truth}    & t = $-7.77^{***}$     \\
    \textit{CBT CoP} VS. \textit{PCT CoP}    & t = $0.69$  \\
    \textit{CBT CoP} VS. \textit{SFBT CoP}    & t = $1.22$  \\
    \textit{CBT CoP} VS. \textit{naive}    & t = $3.32^{***}$  \\
    \textit{CBT CoP} VS. \textit{ChatGPT}    & t = $3.92^{***}$  \\
    \midrule
    \textit{PCT CoP} VS \textit{ground truth}    & t = $-7.90^{***}$     \\
    \textit{PCT CoP} VS. \textit{SFBT CoP}    & t = $0.51$  \\
    \textit{PCT CoP} VS. \textit{naive}    & t = $2.81^{***}$  \\
    \textit{PCT CoP} VS. \textit{ChatGPT}    & t = $3.34^{***}$  \\
    \midrule
    \textit{SFBT CoP} VS \textit{ground truth}    & t = $-8.55^{***}$     \\
    \textit{SFBT CoP} VS. \textit{naive}    & t = $2.13^{**}$  \\
    \textit{SFBT CoP} VS. \textit{ChatGPT}    & t = $2.86^{***}$  \\
    \midrule
    \textit{naive} VS \textit{ground truth}    & t = $-10.52^{***}$     \\
    \textit{naive} VS. \textit{ChatGPT}    & t = $0.90$  \\
    \midrule
    \textit{ChatGPT} VS \textit{ground truth}    & t = $-10.53^{***}$     \\
    \bottomrule
    \end{tabular}
    \caption{Significant test result of human evaluation. $(^{***})p<0.01$, $(^{**})p<0.05$, $(^*)p<0.1$}
    \label{tab:ttest}
\end{table}

\subsection{Detailed Empathy Analysis Results}
\label{apd:empathy}

We also provide the average scores rated by GPT4 on the three dimensions of empathy in Table \ref{tab:gpt4score-origin} for reference. 
\begin{table}[htbp]
    \centering
    \small
\begin{tabular}{lcccc}
\toprule
\textbf{}             &ER       &IP     &EX     &Average \\ 
\midrule
\textit{ChatGPT}      &2.1755   &2.1372 &1.7105 &2.0077\\
\textit{naive}        &1.3051   &1.5498 &1.7975 &1.5508\\
\midrule
\textit{PsyMix}  &1.5676   &1.8831 &2.1169 &1.8558\\
\textit{PCT CoP}      &1.5205   &1.9414 &2.2570 &1.9063\\ 
\textit{CBT CoP}      &1.5187   &1.9006 &2.2558 &1.8917\\
\textit{SFBT CoP}     &1.5266   &1.8705 &2.2355 &1.8775\\ 
\midrule
\textit{ground truth} &1.5501   &1.9316 &2.3095 &1.9304\\
\bottomrule
\end{tabular}%

\caption{GPT4-rated average scores for the answers produced by various counselor chatbots. ER refers to Emotional Reaction, IP refers to Interpretation, and EX refers to Exploration.}
\label{tab:gpt4score-origin}
\end{table}

\section{Examples of Chain-of-Psychotherapies}
\begin{figure}[th]
	\centering
	\includegraphics[width=\linewidth]{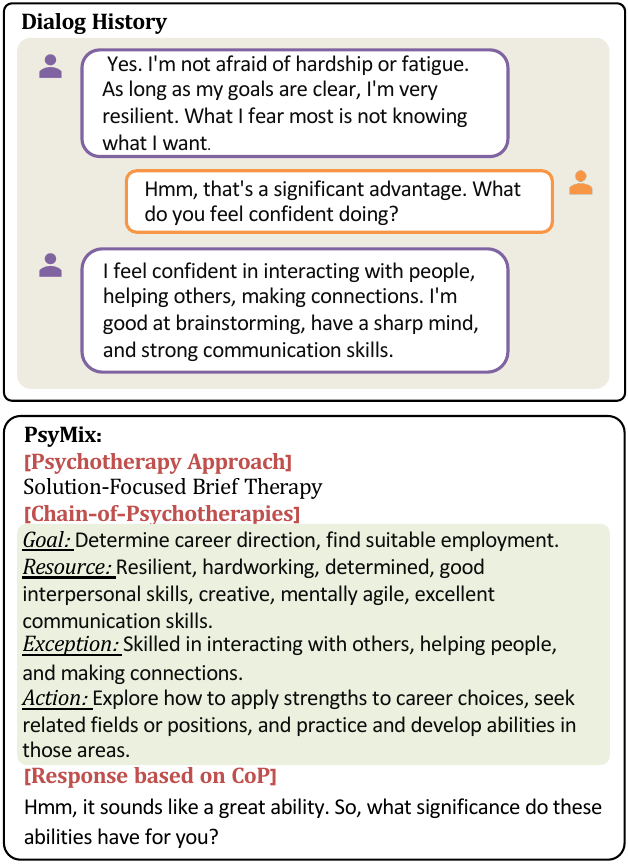}
	\caption{Chain-of-Psychotherapies for case 1.}
	\label{fig:case1}
\end{figure}

\begin{figure}[th]
	\centering
	\includegraphics[width=\linewidth]{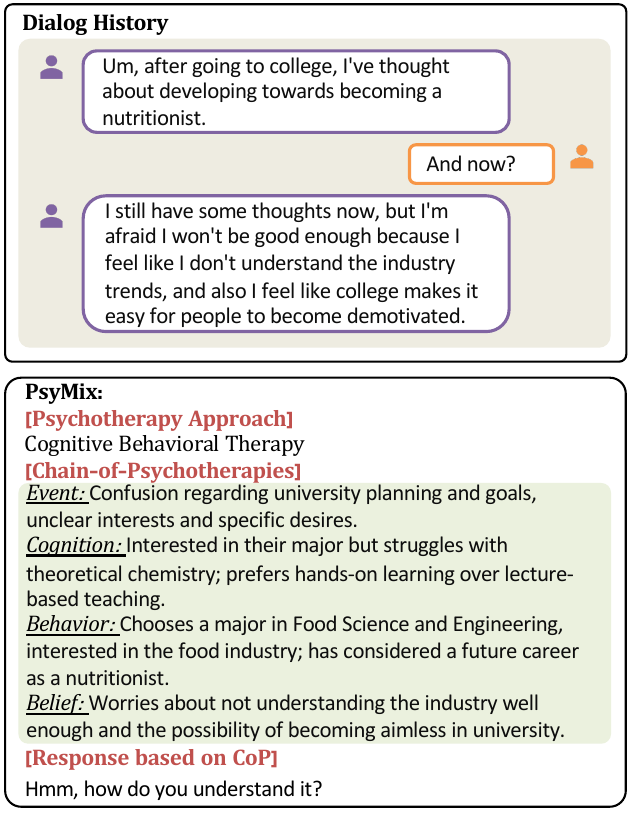}
	\caption{Chain-of-Psychotherapies for case 2.}
	\label{fig:case2}
\end{figure}

In Figures \ref{fig:case1}, \ref{fig:case2}, and \ref{fig:case3}, we present the Chain-of-Psychotherapies procedure for three cases mentioned in Appendix \ref{apd:case}. As shown in the figures, \textit{PsyMix} selects a psychotherapy approach based on the dialog history and generates corresponding analysis before referencing the analysis results to produce a response. 

Through this process, we enable the model to "think" by deeply analyzing the help seeker's situation before generating a response. The resulting responses exhibit higher empathetic quality, addressing the issue of LLM's tendency to generate superficial empathy or quick-fix solutions. Additionally, since we provided analysis results from multiple approaches during the fine-tuning of the model, \textit{PsyMix} integrates the strengths of various psychotherapies.

\begin{figure}[th]
	\centering
	\includegraphics[width=\linewidth]{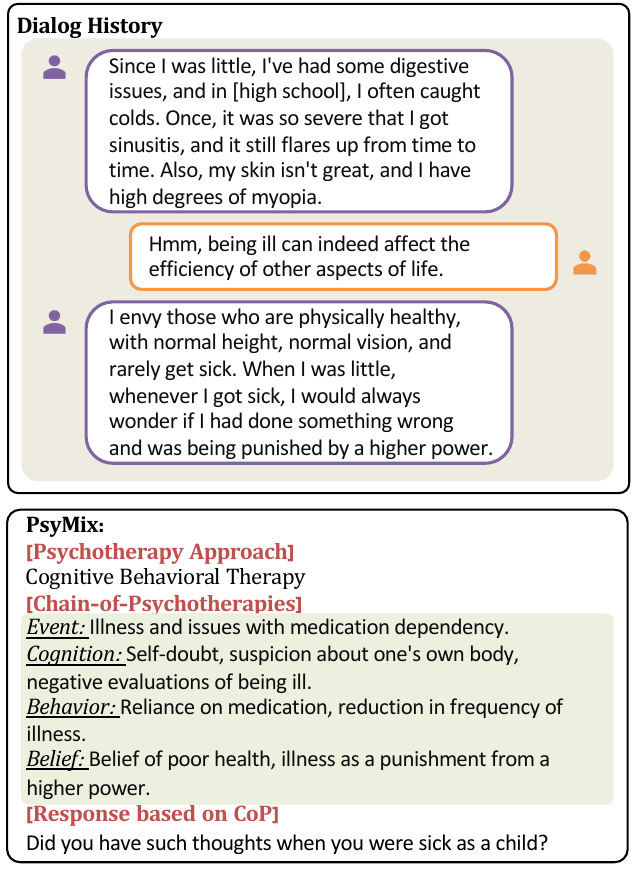}
	\caption{Chain-of-Psychotherapies for case 3.}
	\label{fig:case3}
\end{figure}

\section{Detailed Prompts}

\subsection{Psychotherapy CoT Prompts}
\label{apd:thought_prompt}
Here are the prompts mentioned in Section \ref{sec:cot_generation} for generating psychotherapy analysis of seeker's situation. For each turn, we concurrently focus on three different psychotherapy approaches and generate content tailored to the characteristics of each approach separately.

\begin{tcolorbox}[title = {Prompt for CBT CoP}]
\small
    You are an experienced psychologist. Now, I will provide you with a history of a psychological counseling dialogue. Please analyze the seeker's situation from the perspective of  Cognitive Behavioural Therapy, focusing mainly on the seeker's last statement. Please strictly follow the format below and keep it as concise as possible.
    \\
    
    [Cognitive Behavioural Therapy Analysis]

    *Event: <text>

    *Cognition: <text>

    *Behavior: <text>

    *Belief: <text>
\end{tcolorbox}

\begin{tcolorbox}[title = {Prompt for PCT CoP}]
\small
    You are an experienced psychologist. Now, I will provide you with a history of a psychological counseling dialogue. Please analyze the seeker's situation from the perspective of Person-Centered Therapy, focusing mainly on the seeker's last statement. Please strictly follow the format below.
    \\
    
    [Person-Centered Therapy Analysis]

    *Emotion: <text>

    *Self-Awareness: <text>
\end{tcolorbox}

\begin{tcolorbox}[title = {Prompt for SFBT CoP}]
\small
    You are an experienced psychologist. Now, I will provide you with a history of a psychological counseling dialogue. Please analyze the seeker's situation from the perspective of Solution-Focused Brief Therapy, focusing mainly on the seeker's last statement. Please strictly follow the format below and keep it as concise as possible.
    \\
    
    [Solution-Focused Brief Therapy Analysis]
    
    Seeker's State Assessment:
    
    *Goal: <Text>
    
    *Resource: <Text>
    
    *Exception: <Text>
    
    *Action: <Text>
\end{tcolorbox}


\subsection{\textit{ChatGPT} Baseline Prompt}
\label{apd:baseline_prompt}
The following is the prompt we used for the \textit{ChatGPT} baseline for response generation.

\begin{tcolorbox}[title = {Prompt for \textit{ChatGPT} baseline}]
\small
    Please generate a response to the seeker's last sentence based on the context of the conversation.
    
    Here is the context:
    <context>
    \\
    
    Please respond to the seeker's last sentence coherently and smoothly as a counselor, maintaining a gentle attitude, avoiding repetition of previous remarks, and keeping it concise. Please strictly adhere to the following format for the output:
    
    counselor: <response>
\end{tcolorbox}

\subsection{Empathy Analysis Prompt}
As mentioned in Section \ref{sec:empathy}, we utilize this empathy analysis prompt to instruct GPT-4 to rate seven sets of responses generated by \textit{PsyMix}, along with all other baseline and ablation models, for each context individually. Within the prompt, we furnish comprehensive rating instructions for three dimensions of empathy and the preceding dialogue history.
\begin{figure*}
\begin{tcolorbox}[title = {Prompt for Empathy Analysis}]
\small
You are an expert in psychology. I will provide you with a history of a psychological counseling dialogue and need you to evaluate the empathetic ability of the counselor portrayed in generating responses.

Here are the scoring criteria. The evaluation of empathetic ability will be scored around three dimensions: emotional feedback, understanding, and exploration. Each dimension is set to a score of 1-3, where 1 represents the worst and 3 represents the best. Different responses can have the same scores, but there should be differentiation as much as possible.
\\

\textbf{Emotional Feedback} mainly reflects the warmth, sympathy, and concern expressed in the counselor's replies.
\begin{itemize}
    \item 1 point: No emotional feedback provided.
    \item 2 points: Expresses support but does not explicitly indicate emotions (e.g., everything will get better).
    \item 3 points: Shows empathy towards the seeker, specifically indicating emotions (e.g., I feel sorry for you).
\end{itemize}

\medskip

\textbf{Understanding} refers to the counselor inferring the seeker's feelings and experiences and expressing understanding.
\begin{itemize}
    \item 1 point: No expression of understanding.
    \item 2 points: Expresses understanding but without specific content (e.g., I understand how you feel).
    \item 3 points: Accurately and specifically indicates inferred content (e.g., you must have been very sad at that time) or shares similar experiences (e.g., I sometimes feel very anxious too).
\end{itemize}

\textbf{Exploration} refers to the counselor expressing interest in the seeker's experiences and feelings and gently probing.
\begin{itemize}
    \item 1 point: No interest expressed in the seeker's reply.
    \item 2 points: Expresses interest but in a general manner (e.g., what happened?).
    \item 3 points: Expresses a specific desire to explore some aspect of the seeker's experience (e.g., do you feel lonely now?).
\end{itemize}
Taking into account emotional feedback, understanding, and exploration, please score the response and explain the reasons. The output format is:
\ \\
Scoring Reasons: [Reasons]; 

Emotional Feedback: [Score]; 

Understanding: [Score]; 

Exploration: [Score]; 
\end{tcolorbox}

\end{figure*}

\end{document}